%% file: main.tex
\documentclass{article}
\pdfoutput=1
\usepackage[
  margin=0.8in,
  headheight=12pt,
  headsep=25pt,
  includefoot,
  footskip=30pt,
]{geometry}
\usepackage{graphicx}
\usepackage{booktabs}
\usepackage{tabularx}
\usepackage{geometry}
\usepackage{colortbl}
\usepackage{xcolor}
\usepackage{authblk}
\usepackage{amsmath}
\usepackage{algorithmic}
\usepackage{algorithm}
\usepackage{amsthm}
\usepackage{amssymb}
\usepackage{amsmath}
\usepackage{amsfonts}
\usepackage{tabularx}
\usepackage{multirow}
\usepackage{xcolor}
\usepackage{graphicx}
\usepackage{wrapfig}
\usepackage{caption}
\usepackage{subcaption}
\usepackage{booktabs}
\usepackage{makecell}
\usepackage[utf8]{inputenc} % allow utf-8 input
\usepackage[T1]{fontenc}    % use 8-bit T1 fonts
\usepackage[colorlinks=true,citecolor=OliveGreen]{hyperref}       % hyperlinks
\usepackage{nicefrac}       % compact symbols for 1/2, etc.
\usepackage{microtype}      % microtypography
\usepackage[most]{tcolorbox}
\usepackage[bottom]{footmisc}
\usepackage{tikz}
\usepackage{inconsolata}

\usepackage{natbib}

\input{imports/packages_iclr}
\input{imports/colors}
\input{imports/counter}

\input{math/commands}

\input{math/format}

\input{math/shorthands}
\pdfoutput=1

\tcbset{
  aibox/.style={
    width=\textwidth,
    top=0pt, bottom=0pt, left=5pt, right=5pt,
    colback=white,
    colframe=black,
    colbacktitle=black,
    enhanced,
    center,
    attach boxed title to top left={yshift=-0.1in,xshift=0.15in},
    boxed title style={boxrule=0pt,colframe=white,},
  }
}
\newtcolorbox{AIbox}[2][]{aibox,title=#2,#1}

\definecolor{commentcolour}{rgb}{0.3,0.7,0.2}
\definecolor{backcolour}{rgb}{0.98,0.98,0.98}

\lstdefinelanguage{markdown}{
    comment=[l]{\#},
    morestring=[s]{```}{```},
    commentstyle=\color{commentcolour}\bfseries,
    stringstyle=\color{blue},
    basicstyle=\scriptsize\ttfamily,
    showstringspaces=false,
    breaklines=true,
    breakautoindent=false,
    breakindent=0pt,
    backgroundcolor=\color{backcolour},
}
\lstdefinestyle{mystyle}{
    morekeywords={self},
    basicstyle=\footnotesize\ttfamily,
    keywordstyle=\color{blue},
    commentstyle=\color{commentcolour}\bfseries,
    breaklines=true,
    breakautoindent=false,
    showstringspaces=false,
    stringstyle=\color{red},
    escapechar=@
}
\lstdefinelanguage{PythonPlus}[]{Python}{
  morekeywords=[1]{,as,assert,nonlocal,with,yield,self,True,False,None,} % Python builtin
  morekeywords=[2]{,__init__,__add__,__mul__,__div__,__sub__,__call__,__getitem__,__setitem__,__eq__,__ne__,__nonzero__,__rmul__,__radd__,__repr__,__str__,__get__,__truediv__,__pow__,__name__,__future__,__all__,}, % magic methods
  morekeywords=[3]{,object,type,isinstance,copy,deepcopy,zip,enumerate,reversed,list,set,len,dict,tuple,range,xrange,append,execfile,real,imag,reduce,str,repr,}, % common functions
  morekeywords=[4]{,Exception,NameError,IndexError,SyntaxError,TypeError,ValueError,OverflowError,ZeroDivisionError,}, % errors
  morekeywords=[5]{,ode,fsolve,sqrt,exp,sin,cos,arctan,arctan2,arccos,pi, array,norm,solve,dot,arange,isscalar,max,sum,flatten,shape,reshape,find,any,all,abs,plot,linspace,legend,quad,polyval,polyfit,hstack,concatenate,vstack,column_stack,empty,zeros,ones,rand,vander,grid,pcolor,eig,eigs,eigvals,svd,qr,tan,det,logspace,roll,min,mean,cumsum,cumprod,diff,vectorize,lstsq,cla,eye,xlabel,ylabel,squeeze,}, % numpy / math
}

\title{\texttt{TinyGSM}: achieving $>80\%$ on GSM8k with small language models}

\author{Bingbin Liu}
\affil[1]{Carnegie Mellon University}
\author[2]{Sebastien Bubeck}
\author[2]{Ronen Eldan}
\author[2]{Janardhan Kulkarni}
\author[2]{\authorcr Yuanzhi Li}
\author[2]{Anh Nguyen}
\author[2]{Rachel Ward}
\author[2]{Yi Zhang}
\affil[2]{Microsoft Research}

\date{\vspace{-2em}}

\begin{document}

\maketitle

\begin{abstract}
\noindent
    Small-scale models offer various computational advantages, and yet to which extent size is critical for problem-solving abilities remains an open question. Specifically for solving grade school math, the smallest model size so far required to break the 80\% barrier on the GSM8K benchmark remains to be 34B.
    Our work studies how high-quality datasets may be the key for small language models to acquire mathematical reasoning. We introduce \texttt{TinyGSM}, a synthetic dataset of 12.3M grade school math problems paired with Python solutions, generated fully by GPT-3.5. After finetuning on \texttt{TinyGSM}, we find that a duo of a 1.3B generation model and a 1.3B verifier model can achieve 81.5\% accuracy, outperforming existing models that are orders of magnitude larger.
    This also rivals the performance of the GPT-3.5 ``teacher'' model (77.4\%), from which our model's training data is generated.
    Our approach is simple and has two key components:
    1) the high-quality dataset \texttt{TinyGSM}, 
    2) the use of a verifier, which selects the final outputs from multiple candidate generations.
\end{abstract}

\input{1.intro}

\input{related_works}

\input{2.data}
\section{Solving grade school math with small language models}
The 1.3B version of our phi-GSM models is able to achieve 81.5\% accuracy on GSM8K, a dataset that remains challenging for small-scale models.
The performance comes from sufficient good quality synthetic data and the use of a verifier, which we describe in this section.
\input{3.1.base_model}
\input{3.2.verifier}

\input{5.robust}

\input{conclusion}

\bibliographystyle{iclr2024}
\bibliography{references}

\newpage
\appendix

\input{appendix_data}

\end{document}

%% file: imports/packages_iclr.tex
\usepackage[utf8]{inputenc}
\usepackage{amsmath}
\usepackage{amssymb}
\usepackage{amsthm}
\usepackage{graphicx}
\usepackage{bm,bbm}

\usepackage{listings}
\usepackage{cleveref}
\usepackage{mleftright}
\usepackage{enumitem}
\usepackage{url}
\usepackage{caption}
\usepackage{subcaption}

\usepackage{natbib}

\usepackage{wrapfig}

%% file: imports/colors.tex
\definecolor{RoyalBlue}{RGB}{0,100,170}
\definecolor{OliveGreen}{RGB}{60, 128, 49}
\definecolor{peach}{rgb}{1, 0.56, 0.56}
\definecolor{midgray}{RGB}{150,150,150}
\definecolor{EasternBlue}{RGB}{37,150,190}
\definecolor{sand}{RGB}{250,150,120}
\definecolor{grass}{RGB}{120, 190, 50}
\definecolor{sky}{RGB}{50,150,250}
\definecolor{Orange}{RGB}{250,150,50}
\definecolor{Cerulean}{RGB}{80,150,220}
\definecolor{Emerald}{RGB}{62,156,94}
\definecolor{Rouge}{RGB}{250,95,95}

\definecolor{ColorDef}{RGB}{80, 180, 150}
\definecolor{RevisionRed}{RGB}{240,35,35}
\definecolor{RevisionBlue}{RGB}{80,180,250}
\definecolor{Confirm}{RGB}{103, 162, 229}
\definecolor{RESULT}{RGB}{250,150,50}
\definecolor{TODO}{RGB}{250,150,50}
\definecolor{Future}{RGB}{198, 121, 235}

\definecolor{myGold}{RGB}{231,141,20}
\definecolor{myBlue}{rgb}{0.19,0.41,.65}
\definecolor{myPurple}{RGB}{175,0,124}
\definecolor{Gred}{RGB}{219, 50, 54}
\definecolor{Ggreen}{RGB}{60, 186, 84}
\definecolor{Gblue}{RGB}{72, 133, 237}
\definecolor{Gyellow}{RGB}{247, 178, 16}

%% file: imports/counter.tex
\usepackage{environ}

\newif\ifsolution
\solutiontrue
\newcounter{solnequation}
\newcounter{solneqtmp}
\NewEnviron{soln}{
    \setcounter{solneqtmp}{\value{equation}}
    \setcounter{equation}{\value{solnequation}}
    
    \ifsolution\color{red}\expandafter\textbf{Solution} \BODY\fi%
    \setcounter{solnequation}{\value{equation}}
    \setcounter{equation}{\value{solneqtmp}}
    }

%% file: math/commands.tex
\newtheorem*{namedtheorem}{\theoremname}
\newcommand{\theoremname}{testing}

\newtheorem*{theorem*}{Theorem}

\newtheorem*{question*}{Question}

\theoremstyle{definition}

\newtheorem*{definition*}{Definition}

\newtheorem*{remark*}{Remark}

\theoremstyle{plain}

%% file: math/format.tex
\def\eqref#1{equation~\ref{#1}}
\def\1{\bm{1}}

\DeclareMathAlphabet{\mathsfit}{\encodingdefault}{\sfdefault}{m}{sl}
\SetMathAlphabet{\mathsfit}{bold}{\encodingdefault}{\sfdefault}{bx}{n}

%% file: 1.intro.tex
\section{Introduction}
\label{sec:intro}

\begin{figure}[t]
    \centering
    \includegraphics[width=0.8\textwidth]{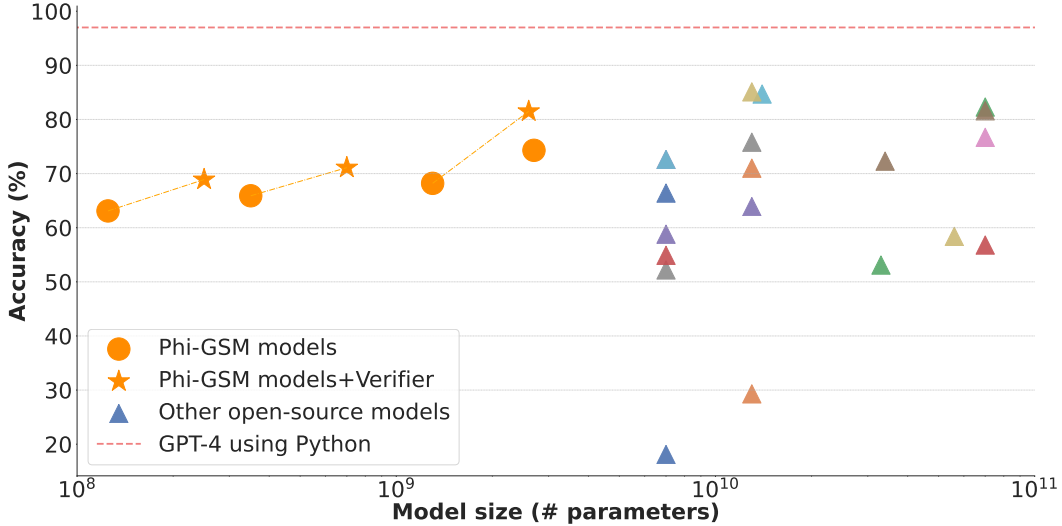}
    \caption{Our results on GSM8K. Please refer to~Table~\ref{tab:comparison} for details.}
    \label{fig:GSM8K}
\end{figure}

One fascinating phenomenon regarding large language models (LLMs) is the emergence of capbilities as both the model and dataset sizes scale up~\citep{wei2022emergent,chan2022emergent}. Among many capabilities, mathematical reasoning is one particular aspect that has received tremendous attention~\cite{lewkowycz2022solving,lightman2023lets}. However, it is unclear to what extend scale is a necessity for mathematical reasoning, and the potential of small language models (SLMs) remains largely under-explored. 

In this work, we push the boundaries of SLMs' math reasoning capabilities. As a first step towards general mathematics, our primary testing ground is grade school math problems, to solve which require both mathematical understanding and language comprehension. The gold-standard benchmark in this regime is GSM8K~\citep{gsm8k}, a collection of 8.8K grade-school math word problems (with a 7k-1k train-test split) that involve $2$ to $11$ reasoning steps. GSM8K has been widely regarded to be challenging for LLMs. Even though the questions appear rather simple for humans, there have been few models that achieve $>80\%$, and they are commonly of prohibitive sizes, e.g. $34$B and above (see Table~\ref{tab:comparison}).

Our goal is to break the $80\%$ barrier on GSM8K while keeping the model size friendly. As previous work shows~\citep{textbooks,textbooks2,eldan2023tinystories}, training data quality is one of the most important factors for enhancing performance of small models. In particular, prompt-engineered synthetic data generation from gigantic models such as GPT-3.5/4 enjoys the clear advantage of desirable data hygiene and controllable diversity. This constituents a teacher-student scenario where the student learns from teacher's generations. On the tasks that the model model already excels at, their guided generations remain one of the highest quality data one can collect for training significantly smaller student models. It is also understood that the student model's performance likely ends up inferior than the teacher, and may fall far short especially when the student is considerably smaller than the teacher~\citep{mirzadeh2019improved,gudibande2023false} \textemdash after all, the teacher places an information-theoretic bottleneck on the student. 

To our surprise, in the case of GSM8K, we are able to bridge the performance gap between the student and teacher, by utilizing a tiny amount of labeled \emph{real} data (the original GSM8K training set of 7k questions) to train an independent verifier model. At test time, the verifier score and select among multiple candidate answers generated from the student, and then we output the highest score generation as the final submission. Note the idea of using a verifier is proposed by the seminal GSM8K paper~\citep{gsm8k}, and here we demonstrate its power of bridging the teacher-student gap, and we conduct a more thorough examination of factors affecting its efficacy.

The contributions of our work are the following:
\vspace{-0.5em}
\begin{itemize}[leftmargin=1.5em]
    \item We introduce \texttt{TinyGSM}, a synthetic dataset containing GSM8K-style math word problems paired with Python solutions, generated fully by GPT-3.5-turbo. TinyGSM consists of 12.3M questions which amount to 1.8B tokens. 
    We demonstrate \texttt{TinyGSM}'s high-quality by finetuning the Phi-1.5 1.3B model (before the use of verifiers) which improves its accuracy from 44.6\% to \textbf{68.2\%} on the GSM8K test set. Notably, our smallest 125M model can also achieve \textbf{63.1\%} after finetuning on \texttt{TinyGSM}.
    
    \item We demonstrate the power of \textit{verifiers} on small-scale models.
    When integrated with a verifier for scoring generations, our models, named Phi-GSM models, achieve performance on par with other open source models that are orders of magnitude larger.
    In particular, our 1.3B model achieves 81.5\% accuracy on GSM8K, as shown in~\Cref{fig:GSM8K}.
    This marks a new state-of-the-arts on billion-parameter-scale models, significantly outperforming existing open-source models and even rivaling the 77.4\% accuracy of GPT-3.5, from which \texttt{TinyGSM} is generated.
    For verifier training, we identify \textit{data diversity} as a crucial element for a verifier's success, and find that the \textit{scaling of the verifier} may be more effective than scaling of the generator: while scaling up from a 125M generator to a 1.3B generator only gives a 5.1\% increase in performance (\Cref{tab:comparison}), scaling up the verifier from 125M to 1.3B leads to a 7.2\% performance boost (\Cref{fig:verifier_sameSize}).
\end{itemize}

\begin{table}
\begin{center}
\small
\begin{tabular}{llrccc}
\Xhline{4\arrayrulewidth}
Model & Base model & Model size & Answer format & Eval method & GSM8K (\%)  \\
\Xhline{3\arrayrulewidth}
\multirow{4}{*}{Llama-2~\citep{llama2}} & \multirow{4}{*}{-}  & 7B &\multirow{3}{*}{nlp} & \multirow{3}{*}{pass@1} &14.6\\
& & 13B &&&  28.7\\
& & 34B &&&  42.2\\
& & 70B &&&  56.8\\
\hline
\multirow{3}{*}{MetaMath~\citep{metamath}} & \multirow{3}{*}{Llama-2} & 7B &\multirow{3}{*}{nlp}&\multirow{3}{*}{pass@1} &  66.5\\
& & 13B &&&  72.3\\
& & 70B &&&  \textbf{82.3}\\
\hline
\multirow{3}{*}{WizardMath~\citep{wizardmath}} & \multirow{3}{*}{Llama-2} & 7B & \multirow{3}{*}{nlp}&\multirow{3}{*}{pass@1} & 54.9\\
& & 13B &&&  63.9\\
& & 70B &&&  \textbf{81.6}\\
\hline
\multirow{4}{*}{MAmmoTH~\citep{mammoth}} & Code-Llama & 7B &  \multirow{3}{*}{code}&\multirow{4}{*}{pass@1} &  59.4\\
& Code-Llama & 12B &&&  64.7\\
& Code-Llama & 34B &&&  72.7\\
& Llama-2 & 70B & nlp &&  76.9\\
\hline
\multirow{2}{*}{Mistral~\citep{jiang2023mistral}} & \multirow{2}{*}{-} &7B & \multirow{2}{*}{nlp} & \multirow{2}{*}{maj1@8} &  52.2\\ 
&  & 8$\times$7B & & &  58.4\\
\hline
\multirow{2}{*}{OVM~\citep{OVM}} & Llama-2 & 7B+7B & \multirow{2}{*}{nlp}& \multirow{2}{*}{verify100@1} & 73.7\\
& Mistral & 7B+7B & &  & \textbf{84.7}\\
\hline
\multirow{2}{*}{Llemma~\citep{llemma}} & \multirow{2}{*}{Llama-2} & 7B & \multirow{2}{*}{nlp}&\multirow{2}{*}{pass@1} & 36.4\\
& & 34B & & & 51.5\\
\hline
\multirow{4}{*}{ToRA-Code~\citep{tora}} & \multirow{4}{*}{Llama-2} & 7B & \multirow{4}{*}{code} & \multirow{4}{*}{COT@1} & 72.6\\
& & 13B & & & 75.8\\
& & 34B & & & \textbf{80.7}\\
& & 70B & & & \textbf{84.3}\\
\hline
\multirow{2}{*}{Orca 2~\citep{mitra2023orca}} & \multirow{2}{*}{Llama-2} & 7B & \multirow{2}{*}{nlp} & \multirow{2}{*}{pass@1} & 55.72\\
& & 13B & & & 65.73\\

\Xhline{3\arrayrulewidth}
Gemini Pro & \multirow{2}{*}{-} & \multirow{2}{*}{-} & \multirow{2}{*}{nlp} & \multirow{2}{*}{maj1@32} & \textbf{86.5}\\
Gemini Ultra~\citep{google2023gemini} & & & & & 94.4\\
\midrule
GPT-3.5-0613 & \multirow{2}{*}{-} & \multirow{2}{*}{-} & \multirow{2}{*}{code} & \multirow{2}{*}{pass@1} & 77.4*\\
GPT-4-0613~\citep{openai2023gpt4} & & & & & \textbf{97.0}*\\
\Xhline{3\arrayrulewidth}
Phi-1.5~\citep{textbooks2} & - & 1.3B & code & \multirow{1}{*}{pass@1} & 44.6\\
\hline
\multirow{4}{*}{Phi-GSM} & Phi-1.5-tiny & 125M & \multirow{4}{*}{code} &  \multirow{4}{*}{pass@1}  & 63.1\\
& Phi-1.5-small & 350M & & & 65.9\\
& Phi-1.5 & 1.3B & & & 68.2\\
& Phi-2 & 2.7B & & & 74.3\\
\hline
\multirow{3}{*}{Phi-GSM+V} & Phi-1.5-tiny & 125M+125M & \multirow{3}{*}{code} &  \multirow{3}{*}{verify48@1}  & 68.9\\
& Phi-1.5-small & 350M+350M & & & 71.3\\
& Phi-1.5 & 1.3B+1.3B & & & \textbf{81.5}\\
\Xhline{4\arrayrulewidth}
\end{tabular}
\end{center}
\caption{Results on GSM8K. * denotes results measured by ourselves. Accuracies \textbf{above 80\%} are labeled in \textbf{bold}. `8$\times$7B' stands for mixture of $8$ experts, and each expert is of 7B parameters. `7B+7B' means a combination of a 7B generation model plus a 7B verifier model. `+V' denotes the use of verifier models.} 
\label{tab:comparison}
\end{table}

%% file: related_works.tex
\section{Related works}

\textbf{Distilling from synthetic data}:
While scaling up has been a useful strategy, it is possible to outpace conventional scaling laws by better use of data~\citep{sorscher2022beyond}.
In the data-scarce case, quality synthetic data serves as an effective workaround~\citep{eldan2023tinystories,textbooks},
and the scaling in dataset size can compensate for a small model size~\citep{edelman2023pareto}.
Additionally, our work uses samples in the true distribution (i.e. the GSM8K train set) differently: given the small dataset size, we believe that the most sample-efficient way to utilize the true train set is to train a verifier---while the 7.4k samples in the GSM8K training set is too small for language model finetuning, it is sufficient for training a good quality verifier that provides 10\% performance boost.
While there have been potential concerns of learning from synthetic data such as loosing diversity or having a drifted distribution mean~\citep{alemohammad2023mad,shumailov2023curse}, \cite{alemohammad2023mad} showed that such degradation can be avoided by including fresh samples from the true distribution during training.

\textbf{Math word problem datasets}
GSM8K~\citep{gsm8k} has been the most common used math word problem dataset for its quality and size.
In comparison, earlier datasets such as MAWPS~\citep{mawps}, ASDiv~\citep{ASDiv} and SVAMP~\citep{SVAMP} are either much smaller in size or of less difficulty.
However, GSM8K questions are too clean to test for robustness.
Motivated by the observation that language models are not robust to the presence of irrelevant context, \cite{GSMIC} proposed GSM-IC (irrelevant context).
Another problem is the GSM8K dataset itself is still too small for training language models.
\citep{ni2023selfsampled} addressed this with self-sampled data.
In a work concurrent to ours, \cite{metamath} bootstraps an original dataset using various augmentation techniques, such as generating multiple answers for the solution, question rephrasing, and backward reasoning.
The proposed MetaMath dataset consists of 40000 questions from GSM8K and MATH~\citep{MATH}.
In comparison, TinyGSM is significantly larger, encompassing 12.3M questions (or equivalently 1.8B tokens).

\textbf{Leveraging multiple generations}:
An important component of our method is to leverage multiple generation.
This idea has been proven successful in many prior works.
A notable example is ``self-consistency''~\citep{wang2022selfconsistency}, which selects the most frequent response among candidates and integrates well with other methods such as progressive-hint prompting~\citep{zheng2023PHP} and model selection~\citep{zhao2023modelSelect}.
However, self-consistency was not particularly helpful in our setup as mentioned in \Cref{subsec:verifier}.
More related to and a direct inspiration of our work is~\cite{gsm8k}, which uses a verifier to select the best response among 100 candidates, leading to an 20\% accuracy boost.
Our work conducts a more thorough study on the design choices of the verifier, including data diversity and the effect of verifier sizes.
Another design choice orthogonal to ours is the supervision signals, such as outcome-based supervision versus process supervision~\citep{,lightman2023lets}.

\textbf{Learning from partial or process supervision}:
In our experiments, we evaluate on the final accuracy only but train on full programs.
Prior work has studied the effect of process versus outcome based supervision.
Process-based supervision is shown to be particularly helpful for complex math problems~\citep{lightman2023lets}, though for general problems one needs to consider a cost-efficacy tradeoff~\citep{uesato2022solving}.
When process supervision is not available, \cite{ni2023selfsampled} proposed to learn from ``self-sampled'' solutions, which allows the model to learn from partially correct self-generated solutions selected based on the execution trace.

\textbf{Self-improvement}:
Several works have explored the idea of ``self-improvement'' where a model evaluates and corrects its own generations, mostly relying on the self-debugging ability of GPT4.
Examples include ``self-refine''~\citep{madaan2023selfrefine} and ``self-verify''~\citep{weng2022verify,zhou2023selfVerify}, both of which ask the model to iteratively provide feedback or verifications on its own generations and refine if needed.
However, such self-improvement abilities have not been discovered in small language models.
This motivated our use of a separate verifier model, which is initialized from the generative model but needs to be fully finetuned for verification.

\textbf{Prompt-based methods}:
\textit{Prompt-based} methods, which find prompts to improve the later conditional generations, are particularly effective for large models.
Examples include \textit{in-context learning}~\citep{GPT3}, where the model learns to perform novel tasks from few-shot examples provided in the prompt,
as well as \textit{Chain-of-Thought}~\citep{CoT}, which shows that explicitly generating intermediate steps can significantly help with reasoning abilities.
However, similar to self-improvements, prompting is targeted at large language models and do not apply for SLMs.

%% file: 2.data.tex
\section{The TinyGSM dataset}
\label{sec:data}

Our objective is to assess the capability of a \textit{small} language model (SLM) on mathematical reasoning.
Ideally, enhancing this mathematical reasoning ability should not compromise the model's competence in language comprehension.
This makes \textit{math word problems}, which necessitate both mathematical and language understanding, a suitable test ground.
We focus on the GSM8K dataset~\citep{gsm8k}, consisting of around 8k grade-school math word problems.
The math concepts in the dataset are elementary and within standard grade-school curricula, but the challenges posed by the natural language problem statement introduce an additional layer of complexity to the task.

\paragraph{\texttt{TinyGSM}: augmenting GSM8K with synthetic generations}

Despite the high quality, the GSM8K training set only contains 7473 problems, which is too small for training a reasonably sized language model~\citep{ni2023learning}.
To alleviate the size issue, we augment the GSM8K training set using GPT-3.5-turbo generated synthetic problems.

We prompt GPT-3.5-turbo to generate problem variants similar to a given question (but not the solution) randomly sampled from the GSM8K training set.
Each problem variant contains both a question and the corresponding solution written in Python, as shown in \Cref{fig:answer_format}.\footnote{Note that the generated problems may be mathematically valid yet violating common sense. For example, some quantities may not be integers.}
Using code allows us to leverage a Python interpreter, circumventing language models' known limitation regarding numerical calculations and code execution.

\begin{figure}
\begin{AIbox}{}
\begin{minipage}[t]{.55\textwidth}
\begin{lstlisting}[language=Python, style=mystyle]
def simple_math_problem() -> int:
    """
    In preparation for her party, Sarah buys 10 trays of food and 8 cases of beverages.
    Each tray costs $50 and each case of beverages costs $20.
    What is the total cost of the trays and beverages?
    """
    trays = 10
    tray_cost = 50
    cases = 8
    case_cost = 20
    tray_total = trays * tray_cost
    case_total = cases * case_cost
    total_cost = tray_total + case_total
    result = total_cost
    return result
\end{lstlisting}
\end{minipage}%
\hfill
\begin{minipage}[t]{.43\textwidth}
\begin{lstlisting}[language=Python, style=mystyle]
def simple_math_problem() -> int:
    """
    Kim has 4 times the number of crayons than 8 less than the number of markers she has.
    If she has 60 crayons, how many markers does she have?
    """
    number_crayons = 60
    number_markers = number_crayons // 4 + 8
    result = number_markers

    return result
\end{lstlisting}
\end{minipage}
\end{AIbox}
\caption{Examples from TinyGSM. The question is given as the docstring of a function, and the solution is the code in the function body.
}
\label{fig:answer_format}
\end{figure}

To enhance robustness, we also generated synthetic problems whose questions contain irrelevant information.
This is achieved by augmenting the GSM-IC dataset~\citep{GSMIC}, which is an augmentation of GSM8K specifically designed to introduce irrelevant context (IC) to the question statement.
These GSM-IC variants constitute to approximately one third of TinyGSM.

The resulting synthetic dataset contains 12.3M problems (i.e. question-solution pairs)
\footnote{This corresponds to 1.8B tokens, which costs around \$3600 to generate according to GPT commercial pricing.} with, 
based on the original 7.4k training set questions and their IC variants.
For each question in the GSM8K train set, the prompt based on this question is shared across API calls, and the source of randomness comes entirely from the generation process.
To encourage diversity, we use temperature sampling and specify in the prompt to encourage the problem variants to be grammatically diverse and contain multiple steps; the exact prompts are provided in~Figure~\ref{fig:tinyGSM_prompt} and in~\Cref{app:prompts}.

\paragraph{Filtering}
To ensure the quality of the synthetic data in \texttt{TinyGSM}, 
we filter out problems that are too short or do not contain numbers,
as well as code solutions which are not executable.
Note that we do not check for the \textit{correctness} of the question or the generated solutions, since the ``ground truth'' solution is not available.
Given the effectiveness of self-consistency~\citep{wang2022selfconsistency},
one might want to filter the problems by keeping the ones which have majority vote only.
We did not adopt this strategy since we find that GPT-3.5-turbo's generations are only consistent on easy problems
\footnote{``Easy'' problems refer to the ones for which a 350M model, trained on a part of TinyGSM, already produces same final answer as GPT-3.5-turbo. 
For example, for an early version of our 350M model, the model only achieves around 50\% on the GSM8K test set, but can achieve more than 87\% on synthetic questions with consistent answers.
In other words, adding more easy problems like these will not help our 350M model bridge the performance gap between itself and GPT-3.5-turbo.},
hence such consistency filtering will remove challenging problems, resulting in a dataset that is too easy to be useful.

\begin{figure}[t]
\begin{AIbox}{}
\begin{lstlisting}[language=Markdown]

Consider the following grade-school math problem: {{question}}

Generate 10 different math problems similar to this math problem.

- Make sure each question uses diverse NLP and includes multiple logical steps.
- After each generated problem, write down a **detailed and complete Python program** to solve the question **step by step** (do NOT give the result directly, **DO NOT write calculations in the comments**).
- The program should contain multiple lines of code and end with 'result = XXX' (Make sure to replace XXX with the actual result of the python program).
- Make sure your Python program is complete and solves the problem. Do **NOT** write things like 'solution to be completed', result = ?, insert your code here etc.
- Give the complete solution to solve the problem, written in Python. Do not write things like 'insert your code here'.
- In each new question, **first end with <|endofquestion|>**, and then start writing the program. Each program should end with <|endofprogram|>.
- Example format: Question X: New question (at least 4 sentences long and use diverse NLP) (without the solution) <|endofquestion|> Complete python code with entire solutions and the correct indent (<|endofprogram|>])
\end{lstlisting}
\end{AIbox}
\caption{The prompt template for generating \texttt{TinyGSM}.}
\label{fig:tinyGSM_prompt}
\end{figure}

%% file: 3.1.base_model.tex
\subsection{Learning from synthetic data}
\label{subsec:base_model}

We finetune the Phi-1.5 125M, 350M and 1.3B models on our \texttt{TinyGSM} from \Cref{sec:data}, and in particular, the 1.3B model reaches \textbf{68.2\%} accuracy.\footnote{The Phi-1.5-small 350M and Phi-1.5-125M variants are pretrained on the same pretraining data as the Phi-1.5 1.3B model.}
\footnote{Performance of training on \texttt{TinyGSM} from scratch is reported in \Cref{tab:noPretrain}.}
We use the Adam optimizer with FP16 during training, with a linear warm-up and a maximum learning rate of 1e-4, a weight decay of 0.01, and an effective batch size of 1024.
The finetuning phase takes up to 20k steps in total. As shown in \Cref{fig:GSM8K}, even without verifiers, our models are already competitive to models of size from 7B and larger.
As an anecdote, an earlier and worse performing version of our Phi-GSM 1.3B model gets 94\% (or 82.5\% from 350M at pass@32, whereas the 750M CodeT5+ model~\citep{wang2023codet5} gets 73.8\% (or 70.5\% from 220M) at pass@100.

%% file: 3.2.verifier.tex
\subsection{Improving small models with a verifier}
\label{subsec:verifier}

While sufficient synthetic data can significantly boost model performance, the performance is still below 70\%.
Does further improvement necessitate larger model and more data then?
There may be two concerns: 
First, there may be a diminishing return in adding extra parameters and data; for instance, while there is a 10\% increase in performance when increasing from around one third of the final size of TinyGSM to two thirds, the final one third of the data provided only marginal gain.
Moreover, even if the small language model is able to fully match the quality of the synthetic data, GPT-3.5-turbo itself can only achieves 77.4\% test accuracy on GSM8K, which seemingly poses a limit on the performance of any models distilling from its generations.

In this section, we show that the use of a verifier can be an effective strategy orthogonal to introducing more and better data, and can even help SLMs exceed the accuracy of GPT-3.5-turbo generations.
The main observation that the best of multiple generations significantly outperforms a single generation.
These generations could be low-temperature generations from different checkpoints of a single run, where taking the best out of generations from 5 checkpoints of (an early version of) our 350M model reaches 75\% accuracy,
similar to findings in temporal ensembling~\citep{laine2016temporal} and snapshot ensembles~\citep{huang2017snapshot}.
\footnote{For utilizing multiple checkpoints, an option is to use model soup~\citep{modelsoup}; however, a uniform soup did not improve the accuracy.
Another option is to perform EMA, which has been shown effective in~\cite{block2023butterfly}.
We found that EMA was not helpful when applied to the 1k-step-interval checkpoints; more frequent averaging is likely required.}
The generations could also be from high-temperature generations based on a single checkpoint;
for instance, the pass@32 accuracy of our 1.3B model is 94\%.

This suggests a promising direction of leveraging multiple generations: we can obtain a great performance boost if we are able to identify the best generation.
This idea is effective yet natural:
The probabilistic nature of the generative process naturally leads to the fact that multiple generations of a language model are more likely to contain a correct solution than a single one.
Empirically, it has been widely observed that pass@$k$ accuracy, namely, the accuracy by taking the best of $k$ generations, is often much higher than pass@1.
The main challenge is that without knowing the labels, the definition of ``best'' is usually unclear.
A workaround is to apply some form of self-selection, such as by choosing the one with the highest logit or the most consistent solution~\citep{wang2022selfconsistency,li2022making}.
There is, however, a notable limitation: generations can be consistent and confident yet inaccurate, making the self-consistency approach through majority voting less effective~\citep{li2022making}.

Given these observations and inspired by findings in~\citep{gsm8k}, we propose to use a separate verifier for selecting candidate generations.
For each base generation SLM, we train a verifier to predict whether a generation is a correct solution to the given question.
During inference, we generate multiple candidate generations using temperature sampling, and select the one with the highest verifier score.

\vspace{-0.3em}
\paragraph{Training data}
The training data consists of the SLM's generations on the labele GSM8K training set questions, paired with the binary labels indicating whether a generation leads to the correct numerical answer. We sample 48 generations for each training set question. The binary label for each generation is based on the final execution result and the ground truth label only, and we do not verify the correctness of intermediate steps. Note that this is the only time where the GSM8K training set is directly utilized in training.

\begin{figure}[H]
\centering
\begin{minipage}{0.5\linewidth}
\begin{tabular}{c|ccccc}
\hline
\multirow{2}{*}{Verfier model size} & \multicolumn{5}{c}{Base generation model size}\\
\cline{2-6}
 & 125M & & 350M & & 1.3B\\
\hline
125M & 68.9 && 68.8 && 71.7 \\
350M & 67.3 && 71.3 && 78.3 \\
1.3B & 76.1 && 79.2 && \textbf{81.5} \\
\hline
\end{tabular}  
\end{minipage}%
\begin{minipage}{0.5\linewidth}
\vspace{0.5cm}
\includegraphics[width=\linewidth]{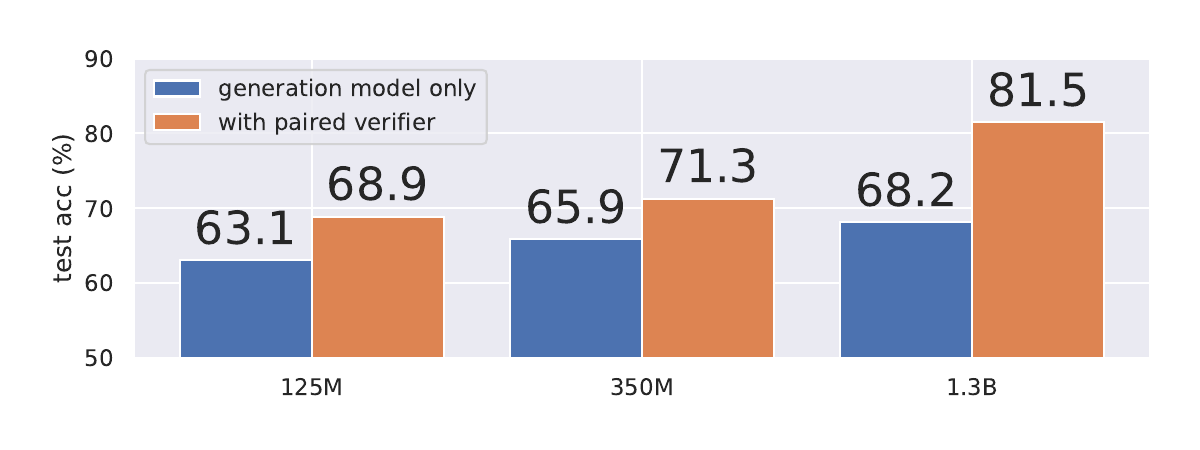}\\
\end{minipage}

\caption{Pass@1 results on GSM8K test set with verifiers. For each test question, we sample $48$ candidate answers from the base generation model, from which we submit the one with highest verifier score as the final answer. The verifier's score on a candidate answer is determined using its score on the last token.}
\label{fig:verifier_sameSize}
\end{figure}

\vspace{-0.3em}
\paragraph{Training setup}
The verifier is trained with a sequence-to-sequence task, where we use the binary label on the entire sequence to supervise each token.
We find this approach improves consistently over training with a sequence classification task (i.e. only predicting a binary label on the entire sequence).
The verifier model is initialized to be the same as the SLM, with an additional prediction head shared across all positions.
All network parameters are updated during verifier training, which significantly outperforms alternatives where only the prediction head is updated, or where the network is trained from scratch.

\vspace{-0.3em}
\paragraph{Checkpoint and data diversity}
The diversity of the training data proves to be important, even if this may result in slightly sacrificing the data quality\footnote{Worse quality data refers to generations from earlier checkpoints, or sampled with a higher temperature: we found that the accuracy for temperature=1 can be more than 10\% worse than temperature=0 (i.e. greedy decoding).}:
we find that including data generated using various temperatures and different checkpoints improves the performance of the verifier.
In particular, the training data for the verifier is generated from checkpoints at 1k, 6k and 12k steps, and both training and testing data use a mixture of data generated with temperature 0.5 and 0.7. Compared to training with generations from a single checkpoint, the use of multiple checkpoints improves the performance from 78.9\% to 81.5\%.

Note also that in general, while we train one verifier for each generative model, verifiers transfer reasonably well across models.
For example, our 1.3B model was able to reach 78.1\% with a 350M verifier trained on generations from a 350M model.

\vspace{-0.3em}
\paragraph{Generation model size vs verifier size} In Figure~\ref{fig:verifier_sameSize}, we present results from a cross-examination of various generation model sizes + verifier model sizes.
Interestingly, while the best accuracy is achieved with configuration with largest sizes, the verifier size seems to play a bigger role than the generation model size.
The effect of model size scaling is surprisingly mild: as shown in~\Cref{tab:comparison}, increasing the base generation model from 125M (Phi-1.5-tiny) to 1.3B (Phi-1.5) only gives a 6\% boost.
On the other hand, the verifier seems to be much more parameter efficient.
For example, 125M generation model + 1.3B verifier can achieve 76.1\%, while 1.3B generation model + 125M verifier gets only 71.7\%~\Cref{fig:verifier_sameSize}.

\begin{figure}
\begin{AIbox}{}
\textbf{Question}:\\
Sophia is thinking of taking a road trip in her car, and would like to know how far she can drive on a single tank of gas. She has traveled 100 miles since last filling her tank, and she needed to put in 4 gallons of gas to fill it up again. The owner's manual for her car says that her tank holds 12 gallons of gas. How many miles can Sophia drive on a single tank of gas?\\\\
\textbf{Generations}:\\
\begin{minipage}[t]{.33\textwidth}
\textbf{1) correct}:
\begin{lstlisting}[language=Python, style=mystyle]
@\textcolor{cyan}{distance\_traveled = 100}@
@\textcolor{cyan}{gas}@@\textcolor{blue}{\_used = 4}@
@\textcolor{blue}{tank\_size = 12}@ 
@\textcolor{blue}{miles\_per}\textcolor{cyan}{\_gallon}\textcolor{blue}{ = distance\_traveled / gas\_used}@
@\textcolor{blue}{result}~\textcolor{cyan}{= miles\_per\_gallon}~\textcolor{blue}{ * tank\_size}@

\end{lstlisting}
\end{minipage}%
\begin{minipage}[t]{.33\textwidth}
\textbf{2) wrong}:
\begin{lstlisting}[language=Python, style=mystyle]
@\textcolor{blue}{distance\_traveled = 100}@
@\textcolor{blue}{gas\_put\_in = 4}@
@\textcolor{blue}{gas\_tank\_size = 12}@
@\textcolor{blue}{result = (distance\_traveled +}@ @\textcolor{cyan}{gas}@@\textcolor{olive}{\_put}@@\textcolor{red}{\_in) / (gas\_tank\_size - gas\_put\_in)}@
\end{lstlisting}
\end{minipage}%
\begin{minipage}[t]{.33\textwidth}
\textbf{3) correct}:
\begin{lstlisting}[language=Python, style=mystyle]
@\textcolor{blue}{miles\_traveled = 100}@
@\textcolor{blue}{gallons\_added = 4}@
@\textcolor{blue}{total\_gallons = 12}@
@\textcolor{cyan}{miles\_per\_gallon = miles}\textcolor{blue}{\_traveled}@ @\textcolor{brown}{// gallons}@@\textcolor{red}{\_added}@
@\textcolor{red}{total\_miles = (total\_gallons - gallons\_added) * miles\_per\_gallon + miles\_traveled}@
@\textcolor{cyan}{result = total\_miles}@
\end{lstlisting}
\end{minipage}
\end{AIbox}
\caption{Visualization of the verifier's token-level predictions. Text colors denote the verifier's prediction scores: \textcolor{blue}{correct}, \textcolor{cyan}{potentially correct}, \textcolor{brown}{potentially wrong}, and \textcolor{red}{wrong}. In all the three examples, the verifier's final prediction (on the last token) aligns with the groundtruth labels. In generation 1) and 2) the verifier's token-level scores appear to be interpretable and aligned with human assessment. However, in generation 3), the scores are rather strange. This suggests the verifier relies on special patterns of the model generations that may not be unversaly generalizable, even though its final predictions are fairly reliable.}
\label{fig:verifier_example}
\end{figure}

%% file: 5.robust.tex
\section{Robustness and decontamination}

\subsection{Contamination test}

While we never use the GSM8K test set during training,
TinyGSM consists entirely of synthetic data generated by GPT models, which may be contaminated since GPT-3.5-turbo may have been exposed to the test set during its own training, which would have led to some generated synthetic samples being replicating part of the test set.
To prevent contamination, we decontaminate TinyGSM by checking for n-gram matches.
We use $n=13$ following standard practices~\citep{GPT3,wei2021finetuned,Du22},
\footnote{n-gram matching is not sufficient for guarding against some other types of contamination (e.g. with respect to paraphrasing).
However, we are not aware of better checks.
One alternative is to check embedding similarity, though our clustering results on CodeGen 350M~\citep{codegen} embeddings suggest that the embedding mostly reflects the semantic (topics) rather than structural/functional similarity,
making it unfit for checking similarity in math questions.
To our knowledge, state-of-the-art papers on training set contamination only test for exact matching~\citep{shi2023detecting,oren2023proving}, and checking for contamination beyond exact match remains an open problem.
}
and remove punctuation and numbers before computing the matching.
Out of the 11.0M unique synthetic questions
\footnote{The number of unique questions is smaller than the number of question-solution pairs since some questions were sampled more than once in the second step of the 2-step generation (\Cref{app:prompts}) and hence have multiple solutions.}
, 22 questions have a nonzero 13-gram match with the test set, and 38k questions (i.e. around 0.35\% of the full set) have non-zero 8-gram matches.
Examples of 13-gram matches are provided in \Cref{app:tinyGSM}.

\subsection{Evaluation on SVAMP}

\begin{figure}[ht]
\centering
\large
\begin{tabular}{c|ccc}
\hline
\multirow{2}{*}{Verfier model size} & \multicolumn{3}{c}{Base generation model size}\\
\cline{2-4}
 & 125M & 350M & 1.3B\\
\hline
125M & 63.2 & 70.0 & 72.2 \\
350M & 64.6 & 68.7 & 72.3 \\
1.3B & 74.1 & 79.0 & 75.6 \\
\hline
\end{tabular}
\caption{SVAMP test accuracies.}
\label{fig:enter-label}
\end{figure}

For evaluating robustness of our models, we test on the SVAMP (Simple Variations on Arithmetic Math word Problems) dataset~\citep{SVAMP}, consisting of 1000 math word problem questions with a focus on arithmetics. SVAMP constructed by applying certain types of variations to a set of base questions. Even though the base questions are generally considered easier than GSM8K
\footnote{See Table 1 in \cite{xie2023decomposition}.}, the variations may often confuse LLMs, thus making it a challenging benchmark for robustness. Our 1.3B model achieves 75.6\% on SVAMP without further finetuning,
indicating the robustness of the model.

%% file: conclusion.tex
\section{Discussions}

In this work, we showed a simple approach that enabled a 1.3B generation model to achieve 81.5\% on the GSM8K dataset, setting a new state-of-the-art for small language models and raising the performance curve for scaling.
Our approach consists of two simple steps:
1) collecting \textit{TinyGSM}, a GPT-3.5 generated synthetic dataset which we will fully release,
and 2) using a verifier that scores how likely a generation is correct, whose quality is boosted by utilizing diverse generations.
Our results provide positive evidence that small language models have more potentials to be unlock and can be used for efficient.
For future directions, 
\begin{itemize}[leftmargin=*]
    \item \textit{Leveraging different formats}: \texttt{TinyGSM} uses Python code as solutions, inspired by the observation that language models tend to struggle at calculations.
    However, we found that different solution formats, i.e. code versus natural language, can be complementary: while code helps circumvent errors related to execution or calculation, it tends to perform worse at questions that require equation solving, likely due to  the fact that the Python syntax does not naturally support equations.
    Properly combining both formats has the potential to further boost performance.
    
    \item \textit{The effect of verifier size}: 
    Our results show that given a budget on the model size, scaling the verifier may be a more efficient use of the parameters.
    This counters our intuition that verification is an easier task than generation (which involves search), though there might be connections to findings in GAN training where the size of discriminator~\citep{arora2018do}.
    Exploring the parameter efficiency in a generation model versus a verifier is an interesting future direction.

\end{itemize}

%% file: appendix_data.tex
\section{Additional details on TinyGSM}

\subsection{Other prompts}
\label{app:prompts}

The majority of the TinyGSM was generated using the prompt in \Cref{fig:tinyGSM_prompt}, where GPT-3.5-turbo is asked to generate both the question and the corresponding solution.
The remaining data, including all data based on GSM-IC, is generated using a two-step process, where the first step prompts the model to generate question variants, and the second step asks to generate Python solutions given a question variant generated in the first step.
The exact prompts are provided in \Cref{fig:tinyGSM_prompt_question}--\Cref{fig:tinyGSM_prompt_code}.

\begin{figure}[H]
\begin{AIbox}{}
\begin{lstlisting}[language=Markdown]
    Design 10 grade-school math problems similar to a given math problem.
    - Make sure the language is different and sufficiently rephrased.
    - Feel free to change the scenario, story and quantities.
    - Make sure the new problems are no easier than the given problem and require more steps to solve.
    - Make sure the new problems are self-complete, clear, unambiguous, and coherent.
    - Please do not include hints or solutions for the new problems.
    
    ## The original math problem
    
    {{question}}
    
    ## New problems
    
    - Problem 1:
\end{lstlisting}
\end{AIbox}
\caption{The prompt template for generating question variants based on GSM8K.}
\label{fig:tinyGSM_prompt_question}
\end{figure}

\begin{figure}[H]
\begin{AIbox}{}
\begin{lstlisting}[language=Markdown]
    Please write 10 questions similar in style to a given question, where the new questions also contain irrelevant information. Make sure to change the name and scenarios.

    # Original question
    
    {{question}}
    
    # New questions
    - Question 1:
\end{lstlisting}
\end{AIbox}
\caption{The prompt template for generating question variants based on GSM8K-IC.}
\label{fig:tinyGSM_prompt_questionIC}
\end{figure}

\begin{figure}[ht]
\begin{AIbox}{}
\begin{lstlisting}[language=Markdown]
    Please generate a detailed and complete Python program to solve a given math question.

    - Use variables to represent the quantities in the question.
    - Solve the question **step by step** (do NOT give the result directly, **DO NOT write calculations in the comments**).
    - The program should contain multiple lines of code and end with 'result = XXX' (Make sure to replace XXX with the actual result of the python program!!!).
    - Then, the result should be printed out in the format of f'<<<{result}>>>'.
    - Make sure your Python program is complete and solves the problem. Do **NOT** write things like 'solution to be completed', result = ?, insert your code here etc.
    - Give the complete solution to solve the problem in Python. Do not write things like 'insert your code here'.
    - You should solely rely on the Python program to solve the question. Do not do calculations in the comments. The comment should not include any numbers.
    - Try to use fewer comments since they are expensive.
    - If you really want to solve equations like x = ..., try to use ``import sympy`` and **sympy.solve()**. sympy.solve(expression) returns the roots of the expression. Do not write down calculations in the comments!
    - If you need to calculate the ceiling of a number, use `import math` then `math.ceil().`
    - End the Python program with <|endofprogram|> in a new line.
    
    ### Question
    
    {{question}}
    
    ### Program
\end{lstlisting}
\end{AIbox}
\caption{The prompt template for generating code solution for a given question.}
\label{fig:tinyGSM_prompt_code}
\end{figure}

\subsection{Contamination check: 13-gram collisions}
\label{app:tinyGSM}

There are 22 questions (out of 11.0M) with 13-gram collisions to test set questions.
Examples are shown in~\Cref{fig:n_grams_examples}.

\begin{figure}[ht]
\begin{AIbox}{}
\begin{lstlisting}[language=Markdown]
    # Q: Daniel has   brothers His older brother is   years older than   times Daniel's age when Daniel was  years old His younger brother is   years old which is  the age of the older brother What is their combined age

    match:
    In a family there are   brothers and   sisters All sisters are the same age which is   One of the brothers is  **years old which is  the age of the older brother What is** the total age of all these siblings

    # Q: Two cars are driving on a highway The first car is traveling at an average speed of   miles per hour while the second car takes a minute break after driving for   minutes how long can they remain stopped before the first car catches up with them

    match:
    **Two cars are driving on a highway The first car is traveling at an average speed of   miles per hour** when the second car passes it at an average speed of  miles per hour  If both cars continue on the highway at the same speed how many miles will separate them after  hours

    # Q: Leo and Nora sold lemonade at a stand Leo sold  cups at  cents each and Nora sold  cups at  cents each They decided to split the money equally How much money did each of them get

    match:
    While walking down the street with his  young siblings Greg found  To be fair to his siblings he **decided to split the money equally How much money did each of them get**

    # Q: A bookstore is selling a book for  while is a  discount from the original price What was the original price of the book
    
    match:
    Kyle bought last year's bestselling book for  This is with a **discount from the original price What was the original price of the book**

\end{lstlisting}
\end{AIbox}
\caption{The prompt template for generating code solution for a given question.}
\label{fig:n_grams_examples}
\end{figure}

\section{Pretrained vs Random Init}
In this section, we present a comparison of training on \texttt{TinyGSM} from a random initialization versus from a pretrained model.
\begin{table}[H]
    \centering
    \begin{tabular}{c|c|c|c}
    \hline
        Model size & 125M & 350M & 1.3B \\\hline
        Random Init & 53.1 & 55.5 & 57.3 \\\hline
        pretrained &  63.3 & 65.9 & 68.2 \\\hline 
    \end{tabular}
    \caption{Comparison of performance with and without pretraining.}
    \label{tab:noPretrain}
\end{table}